\documentclass{article}

\usepackage{microtype}
\usepackage{graphicx}
\usepackage{subfigure}
\usepackage{booktabs} 
\usepackage{amsmath}
\usepackage{amssymb}
\usepackage{graphicx}

\usepackage{latexsym}

\usepackage{hyperref}

\usepackage[accepted]{icml2020}

\newcommand*\entmaxtext{entmax}
\DeclareMathOperator*{\entmax}{\mathsf{\entmaxtext}}
\newcommand*\aentmax[1][\alpha]{\mathop{\mathsf{#1}\textnormal{-}\mathsf{\entmaxtext}}}

\icmltitlerunning{Sparse Graph to Sequence Learning}

\begin{document}

\twocolumn[
\icmltitle{Sparse Graph to Sequence Learning for \\ Vision Conditioned Long Textual Sequence Generation}




\begin{icmlauthorlist}
\icmlauthor{Aditya Mogadala}{to}
\icmlauthor{Marius Mosbach}{to}
\icmlauthor{Dietrich Klakow}{to}
\end{icmlauthorlist}

\icmlaffiliation{to}{Spoken Language Systems, Saarland University, Saarland Informatics Campus, Germany}

\icmlcorrespondingauthor{Aditya Mogadala}{amogadala@lsv.uni-saarland.de}

\icmlkeywords{Graph-to-Sequence Learning, Graph Transformer, Sparse Graph Encoder}

\vskip 0.3in
]



\printAffiliationsAndNotice{}  

\begin{abstract}
Generating longer textual sequences when conditioned on the visual information is an interesting problem to explore. The challenge here proliferate over the standard vision conditioned sentence-level generation (e.g., image or video captioning) as it requires to produce a brief and coherent story describing the visual content. In this paper, we mask this Vision-to-Sequence as Graph-to-Sequence learning problem and approach it with the Transformer architecture.  To be specific, we introduce \textbf{S}parse \textbf{G}raph-to-\textbf{S}equence \textbf{T}ransformer (SGST) for encoding the graph and decoding a sequence. The encoder aims to directly encode graph-level semantics, while the decoder is used to generate longer sequences. Experiments conducted with the benchmark image paragraph dataset show that our proposed achieve 13.3\% improvement on the CIDEr evaluation measure when comparing to the previous state-of-the-art approach.
\end{abstract}

\section{Introduction}
\label{sec:intro}
Most of the methods which address vision conditioned textual sequence generation have concentrated on shorter sequences (e.g., phrase or sentence). Usually, these methods employ a standard encoder-decoder framework~\cite{cholearning:2014,bahdanau:2014}, where the encoder encodes an image into fixed vector representation and then the decoder decodes them into a textual sequence. Several improvements were seen in the recent years over earlier proposed methods where visual features are upgraded with bottom-up~\cite{anderson:2017} encoding, encoder-decoder architecture added with attention~\cite{xu:2015} and training is achieved with reinforcement for sequence decoding~\cite{rennie:2016}. However, most of these methods fail to capture salient objects observed in the image and generate textual sequences which are generic and simple. A possible reason identified~\cite{yaoexploring:2018} is that visually grounded language generation is not end-to-end and largely attributed to the high-level symbolic reasoning. It is also observed that the high-level reasoning is natural for humans as we inherently incorporate \textit{inductive bias} based on common sense or factual knowledge into language~\cite{kennedyspatial:2007}, however, this is ineffective for vision conditioned textual sequence generation due to gap between visual information and language composition. This gap widens more when longer textual sequences need to be generated when conditioned on visual information.

\begin{figure}
    \centering
        \includegraphics[width=0.49\textwidth]{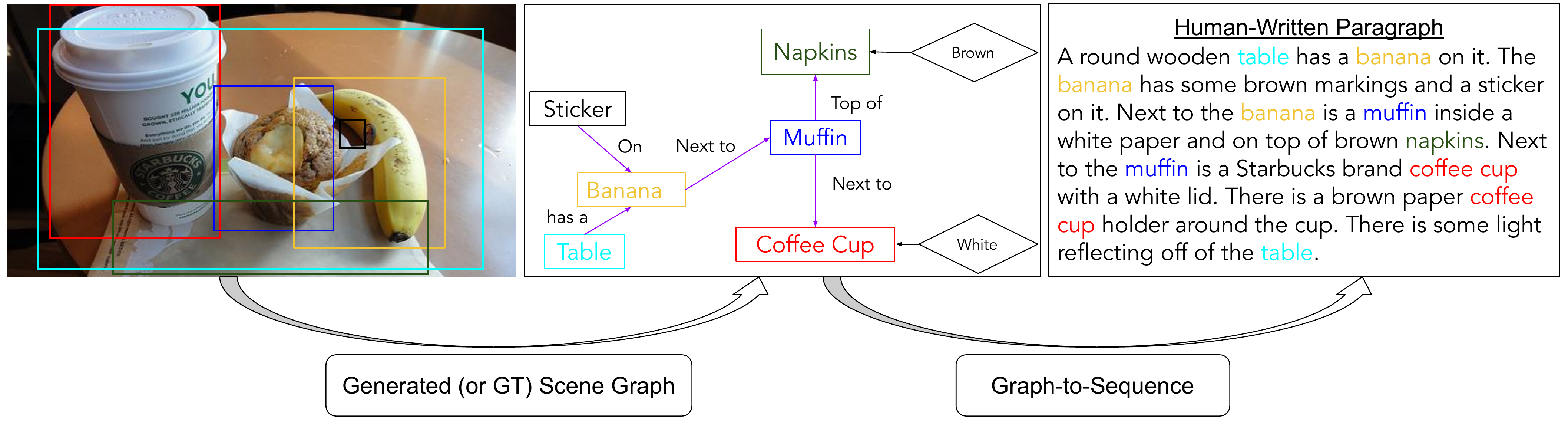}
    \caption{\label{fig:gtosidea}An example of graph-to-sequence from the image-paragraph dataset. Rectangles represent objects, connections between objects represent their relationships and diamond represent the attributes of the objects.}
\end{figure}

In NLP, structured inputs (e.g., graph structures, table data) are omnipresent as a representation of natural language. Recently, several works~\cite{beckgraph:2018} have explored changing them into sequences for different applications~\cite{song:2018,koncel:2019}. Inspired from it, we propose to incorporate graph structure as an inductive bias for vision conditioned textual sequence generation. This is achieved by abstracting visual information (e.g., image) into a scene graph~\cite{johnson:2015} to add complementary strength of symbolic reasoning to multimodal learning.  Scene graphs connect the visual objects, their attributes, and their relationships in an image by directed edges. Figure~\ref{fig:gtosidea} presents the visualization of the overall idea.

However, the major challenge is embedding the scene graph structure into vector representations for seamless integration into the encoder-decoder learning framework. Also, such representation should facilitate the sequence decoder to generate longer sequences. Therefore, in this paper, we introduce \textbf{S}parse \textbf{G}raph-to-\textbf{S}equence \textbf{T}ransformer (SGST) for embedding scene graph by understanding structured sparsity and then decoding it into the textual sequence. This approach builds upon Transformer~\cite{vaswani:2017} encoder-decoder architecture as Transformer based decoders~\cite{radford:2018,yangxlnet:2019} have already shown their ability to decode longer sequences, however, they are less explored in encoding graph structures. Nevertheless, there has been some interest recently~\cite{yungraph:2019,zhugraph:2019}, but, many methods proposed earlier to encode graphs into vector representation are mostly based on Graph Convolutional Networks (GCN)~\cite{kipfsemi:2016}. We hypothize that SGST is a more effective approach for our problem than GCN as it performs global contextualization of each vertex than focused portions in GCN (e.g., adjacent vertices) allowing direct modeling of dependencies between any two nodes without regard to their distance in the input graph. Furthermore, SGST incorporates sparse attention mechanism~\cite{peters:2019} in the self-attention of Transformer architecture allowing it to assign zero probabilities for irrelevant graph vertices or tokens in a sequence. This aids SGST to effectively encode graphs and decode longer sequences.

\section{Vision-to-Sequence as Graph-to-Sequence Learning}
\label{sec:gtoslearning}
A standard approach for vision conditioned language generation is using a vision-language pair $(v,y) \in \mathcal{(V,Y)}$. We first modify it as the graph-language pair i.e., $(g,y) \in \mathcal{(G,Y)}$, where $g=(g_1,g_2,...,g_m)$ is the set of $m$ nodes denoting visual objects, relationships and attributes, while $y=(y_1,y_2,...,y_n)$ is the target description with $n$ tokens. For learning, a language model decoder is conditioned on a graph encoder to learn the parameter $\Theta$ for estimating the conditional probability $P(y|g;\Theta)$ using a log likelihood as the objective function $L(\Theta; \mathcal{(G,Y)}) = \sum_{(g,y)\in \mathcal{\mathcal{(G,Y)}}}\log P(y|g;\Theta)$. 
The conditional probability $P(y|g;\Theta)$ can be factorized according to the chain rule of probability: $P(y|g;\Theta) = \prod_{t=1}^{n} P(y_t | y_{<t}, g; \Theta)$, where $y_{<t}$ denotes the preceding tokens before the index $t$. Furthermore, adding attention mechanism~\cite{bahdanau:2014} between the encoder and decoder will identify the focus of the source representation when predicting the tokens while generation is an additional improvement.

\section{Sparse Graph-to-Sequence Learning}
The sparse Graph-to-Sequence learning is achieved with a sparse Transformer as Graph Encoder and a standard Transformer decoder for sequence generation. 

\subsection{Sparse Graph Transformer as Encoder}
\label{sssec:sparsetransformer}
Our Graph Encoder is inspired by the self-attention use of the Transformer on the sequential data. It can be seen resembling GNN by replacing the token sequence as an unlabeled directed acyclic graph (DAG). To ensure, all scene graphs generated for images are unlabeled DAGs, we replace relations representing labels between vertices by new vertices. Further, the new vertices are connected with the object vertices such that the directionality of the former edge is maintained.  We also introduce a global vertex to connect the entire graph.

Therefore, in the final graph $G = (V,A)$, $V$ embeds objects, relationships, attributes and the global vertex into dense vector representations (combined with their positional encodings), resulting in a matrix $\mathbf{V}^0 =[\mathbf{v}_i], \mathbf{v}_i \in \mathbb{R}^{d}$, given as input to the encoder. $A$ denote the adjacency matrix showing the connection between vertices. Now, each vertex representation $\mathbf{v}_i$ is self-attended with the \textit{sparse graph multi-head attention} over the other neighbourhood vertices to which $v_i$ is connected in the $G$. We use an $N$-headed self attention setup, where $N$ independent attentions are calculated.

The self-attention in the Transformer is densely connected. Given $n$ query contexts and $m$ sequence items under consideration, attention computes, for each query, a weighted representation of the items i.e., \textit{scaled dot-product attention} given in Equation~\ref{eqn:scaledattn}. 
\begin{equation}
\label{eqn:scaledattn}
    \mathsf{Att}(\mathbf{Q}, \mathbf{K}, \mathbf{V}) = \mathsf{softmax}
\left(\frac{\mathbf{Q}\mathbf{K}^\top}{\sqrt{d}}\right) \mathbf{V}
\end{equation}

where $\mathbf{Q} \in \mathbb{R}^{n \times d}$ contains representations of the queries, $\mathbf{K}, \mathbf{V} \in \mathbb{R}^{m \times d}$ are the \textit{keys} and \textit{values} of the items attended over, and $d$ is the dimensionality of these representations. For multiple heads ($\mathsf{H}_i$), $\mathsf{Att}$ is calculated separately.
\begin{equation}
\label{eq:head}%
\mathsf{H}_i(\mathbf{Q}, \mathbf{K},\mathbf{V})\!=\!\mathsf{Att}(\mathbf{QW}_i^Q\!\!,\mathbf{KW}_i^K\!\!,\mathbf{VW}_i^V\!)
\end{equation}

However, we hypothize that adding sparsity to our graph Transformer encoder at different levels is beneficial. There are several advantages to it: (1) eliminate unnecessary vertices that are still taken into consideration to a certain extent for calculation of attention weights~\cite{martins:2016,correia:2019} (2) reduce memory and computational requirements with factorizations of the attention matrix~\cite{child:2019}. We concentrate on the first scenario and present more details in the following.

\paragraph{Sparse Graph Self-Attention} For $\mathcal{N}_i$, the neighborhood of $v_i$ in $G$, we compute the self-attention of a single head using each vertex $v_i$ with vertices $v_j$ in a single-hop using Equation~\ref{eq:attncal}.
\begin{equation}
\label{eq:attncal}
 \mathsf{Att}_G = \sum_{j \in \mathcal{N}_i}\beta_{ij}\mathbf{W}^V\mathbf{v}_j
\end{equation}
where $\mathbf{W}^V \in \mathbb{R}^{d\times d}$ and $\beta_{ij}$ is given by Equation~\ref{eq:attnweight}.
\begin{equation}
\label{eq:attnweight}
\beta_{ij} = \mathsf{Normalize}(\mathbf{v}_i,\mathbf{v}_j)
\end{equation}

Further, to introduce sparse attention, we modify the $\mathsf{Normalize}$ by simply replacing $\mathsf{softmax}$ with $\aentmax$ in the attention heads. That is, $\mathsf{softmax}$ in Equation~\ref{eqn:scaledattn} is modified as follows.
\begin{equation}
\label{eq:alphentmax}
    \mathsf{Normalize}(\mathbf{q}_i, \mathbf{k}_j) = \aentmax (\mathbf{z})
\end{equation}

The $\aentmax(\mathbf{z})$ is given by Equation~\ref{eq:entmaxz}~\cite{blondel:2019} and $\mathbf{z}$ is provided by Equation~\ref{eq:zee}.

\begin{equation}
\label{eq:entmaxz}
\aentmax(\mathbf{z}) = \mathsf{ReLU}[(\alpha - 1){\mathbf{z}} - \tau\mathbf{1}]^{\frac{1}{\alpha-1}}
\end{equation}

\begin{equation}
\label{eq:zee}
 \mathbf{z} = \frac{\mathbf{W}_i^Q\mathbf{q}_i{(\mathbf{W}^K\mathbf{k}_j)}^\top}{\sqrt{d}}
\end{equation}

where $\mathbf{q}_i, \mathbf{k}_j$ are query and key of $v_i$ and $v_j$ respectively. $\tau$ is unique threshold and $\mathbf{1}$ is vector of all ones. For the experiments, following~\cite{peters:2019}, we fixed $\alpha$=1.5 and also used sparse attention with a different learned $\alpha = 1 + \mathsf{sigmoid}(\mathsf{att\_scalar}) \in [1, 2]$ for the each attention head~\cite{correia:2019}. We tie all $\alpha$ values between encoder-decoder and $\mathsf{att\_scalar} \in \mathbb{R} $ is a parameter per attention head.


\paragraph{Encoder} Output ($\mathsf{Att}^N_G$) of $N$ attention heads is concatenated and added to $\mathbf{v}_i$ to attain $\mathbf{\hat{v}}_i$. Further, $\mathbf{\hat{v}}_i$ is passed through different computations in the encoder to transform into $\mathbf{h}_i^{enc}$ given as follows. 
\begin{eqnarray}
    \mathbf{\tilde{v}}_i &=& \mathsf{TransFunction}(\mathsf{LayerNorm}(\mathbf{\hat{v}}_i)) \\
    \mathbf{h}^{enc}_i &=& \mathsf{LayerNorm}(\mathbf{\tilde{v}_i} + \mathsf{LayerNorm}(\mathbf{\hat{v}}_i))
\end{eqnarray}
Where $\mathsf{TransFunction}(\mathbf{x})$ is a two layer feedforward network with a non-linear transformation between layers. To increase the depth of network, blocks are stacked $\mathsf{L}$ times, with the output of layer $l-1$ taken as the input to layer $l$, so that $\mathbf{v}^{l}_i = \mathbf{h}^{enc(l-1)}_i$. Stacking multiple blocks allows information to propagate through the graph.

\subsection{Sparse Sequence Transformer as Decoder}
Our sequence decoder is built on the principle of sequential Transformer decoder. It predicts the next token $y_{t}$ given all the previous tokens $y_{<t} = y_{1},...,y_{t-1}$. Context attention $\mathbf{C}_{att}$ is computed by performing \textit{sparse context attention} for single head over the output ($\mathbf{h}^{enc(l)}_i$) of the graph encoder and decoder hidden state ($\mathbf{h}_t^{dec}$) at each timestep $t$ given as follows.

\begin{equation}
\label{eq:contatt}
 \mathsf{C}_{att} = \sum_{j \in \mathcal{N}_i}\gamma_{j}\mathbf{W}^G\mathbf{h}^{enc(l)}_j
\end{equation}

where $\mathbf{W}^G \in \mathbb{R}^{d\times d}$ and $\gamma_{j}$ is given by Equation~\ref{eq:contattw} and is further modified according to Equation~\ref{eq:alphentmax}.

\begin{equation}
\label{eq:contattw}
\gamma_{j} = \mathsf{Normalize}(\mathbf{h}_t^{dec},\mathbf{h}^{enc(l)}_j)
\end{equation}
We also modify the masked self-attention into masked \textit{sparse sequence self-attention} in the similar manner as \textit{sparse graph self-attention}. Due to space limitation, we present overall architecture in the supplemental material.

\section{Training and Inference}
\label{sec:trainandinfer}
We use Transformer with $\mathsf{L} = 6$ layers and $\mathsf{H} = 8$ heads both for encoder and decoder. To optimize, we use Adam~\cite{kingma:2014} with $\beta$2=0.98. Input embeddings and hidden size is set to 512 and batch size of 2048. The $\mathsf{TransFunction}$ has an intermediate size of 2048. All models are trained between 8 and 12 epochs. During inference, beam search is used with beam size = 5.

\section{Experimental Setup}

\subsection{Dataset and Evaluation Metrics}
To evaluate our proposed approach, we used the image paragraph dataset~\cite{krause:2017} containing images aligned with the textual sequences that are longer than an usual sentence-level caption (e.g., MSCOCO~\cite{lin:2014}) i.e., paragraph. The dataset contains 19,551 image-paragraph pairs. On average, each paragraph has 67.5 words and each sentence in the paragraph consists of 11.91 words. Following the settings of~\cite{krause:2017}, we split the dataset into 14,575 images for training, 2,487 for validation and 2,489 for testing. We evaluated image paragraph generation with the widely used language generation metrics such as BLEU~\cite{papineni:2002}, METEOR~\cite{lavie:2014}, and CIDEr~\cite{vedantam:2015}. 

\subsection{Image Scene Graph Generation}
\label{ssec:imsgg}
Since all images from the image-paragraph dataset are part of the Visual Genome (VG) dataset~\cite{krishna:2017}. We could have directly used the ground truth (GT) scene graph annotations, i.e., objects and their pairwise relationships along with attributes. However, to show that our method can be applied to any image, we generated a scene graph for the images using a trained model that can do object proposal detection (to detect and classify objects), relationship classification (classify relationships between objects), and the attribute classification. To overcome the noisy annotations present in the dataset, similar to~\cite{yangauto:2019} we filter and keep $305$ objects, $103$ attributes, and $64$ relationships to train our detector and classifiers.

To be specific, to train the object detector we used Faster-RCNN~\cite{ren:2015} for extracting 36 RoI features in similar manner as~\cite{anderson:2017}. The detected objects are further used as the input to the relationship classifier~\cite{zellers:2018} to predict a relationship between two objects and also attribute classifier to attain top-3 attributes per object. Now, for each image using the predicted objects, relationships, and attributes, an image scene graph can be built.

\section{Results and Discussion}

Table~\ref{ipg:results} presents the automatic metrics comparison, while Table~\ref{ipg:langanal} shows the language analysis performed to know the choice of the vocabulary used by our models in paragraph generation. From the automatic metrics presented in the Table~\ref{ipg:results} we observe that the BLEU score which is dependent on word overlap has only seen an improvement of around 8\% when SGST ($\aentmax$) (w/ GT scene graphs) compared against~\cite{wangconvolutional:2019}, while CIDEr that cares about overall semantics of the generated paragraph has a gain of 13.3\%. Similarly, when we try to analyze the impact of Table~\ref{ipg:langanal}, we observe that it depicts the usage of grammar in the generated paragraphs. In contrast to single sentence generation, the paragraphs should have a smooth transition between sentences having high coherence. In general, pronouns capture such a transition well while verbs provide the actions observed in the scene. We observe that our proposed models produce more verbs and pronouns while generating concise paragraphs (i.e., minimal average length).

\begin{figure}
    \centering
        \includegraphics[width=0.45\textwidth]{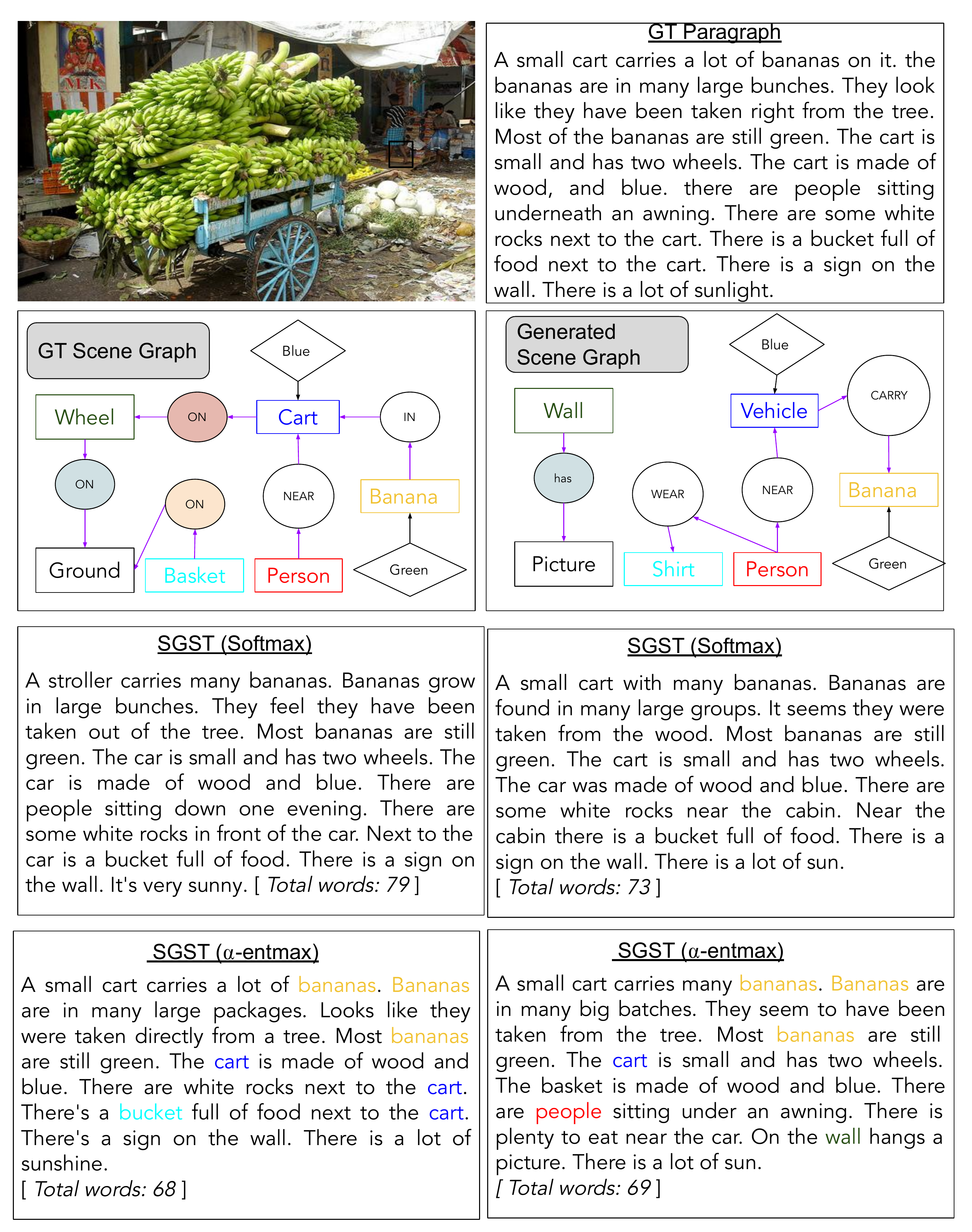}
    \caption{\label{fig:allres} Qualitative results of generated paragraphs (only partial graph is shown w/o global vertex).}
\end{figure}

Figure~\ref{fig:allres} shows an example paragraph generated by the baseline SGST ($\mathsf{softmax}$) and best approach SGST ($\aentmax$) using both GT and the generated scene graph. It is interesting to observe that paragraph generated by both models are highly relevant to the image. However, SGST ($\aentmax$) generated a more coherent and brief paragraph with lesser words.

\begin{table}[!ht]
\small
\centering

\begin{tabular}{lccc}
\toprule
Method                                             & C & M & B-4 \\ 
\midrule
\cite{karpathy:2015}            & 11.06  & 12.82   & 7.71   \\
\cite{krause:2017}                & 13.52  & 15.95  & 8.69   \\
\cite{liang:2017}                    & 20.36   & 18.39  & 9.21   \\
\cite{chatterjee:2018} & 20.93   & 18.62   & 9.43   \\
\cite{wangconvolutional:2019} & 25.15 & 18.82  & 9.67   \\
\midrule
w/ generated scene graphs & &  &  \\
\midrule
SGST ($\mathsf{softmax}$) & 25.22  & 18.95 & 9.84   \\
SGST (1.5-$\entmax$) & 25.46 & 19.01  & 9.86 \\
SGST ($\aentmax$) & 26.01 & 19.16  & 10.05  \\
\midrule
w/ GT scene graphs &  &  &    \\
\midrule
SGST ($\mathsf{softmax}$) & 26.89 & 19.16 & 10.03   \\
SGST (1.5-$\entmax$) & 27.51 & 19.20  & 10.13   \\
SGST ($\aentmax$) & \textbf{28.50} & \textbf{19.25} & \textbf{10.45}  \\
\bottomrule
\end{tabular}
\caption{Performance of our proposed methods in comparison with other state-of-the-art using CIDEr (C), METEOR (M), and BLEU-4 (B-4) measures on the image paragraph dataset. All values are reported as percentage (\%).}
\label{ipg:results}
\end{table}

\begin{table}[!ht]
\small
\centering
\begin{tabular}{lccccc}
\toprule
Method  & AvgLen & StdDev & N & V & P \\
        & (words) &  (words) &  &  & \\
\midrule
Baseline & 70.47  & 17.67  & 24.77 & 13.53 & 2.13   \\
\midrule
w/ generated & &  & & & \\
\midrule
($\mathsf{softmax}$) & 57.33  & 14.01  & 26.00 & 14.33  & 3.67   \\
(1.5-$\entmax$) & 59.66  & 18.71 & 27.00 & 15.00 & 3.33 \\
($\aentmax$) & 54.66  & 11.71  & 25.33 & 15.00 & 3.33 \\
\midrule
w/ GT  &  &  &  & &  \\
\midrule
($\mathsf{softmax}$) & 62.00  & 14.93  & 27.00 & 15.67  & 4.00 \\
(1.5-$\entmax$) & 61.33  & 16.19  & 27.33 & 14.67  &  4.00 \\
($\aentmax$)& 56.33  & 11.15  & 25.67 & 15.67 &  3.00 \\
\midrule
Human & 67.51  & 25.95  & 25.91 & 14.57 & 2.42  \\
\bottomrule
\end{tabular}
\caption{Language Analysis is performed to comprehend the choice of vocab used by our models in generation. Regions-Hierarchical model from~\cite{krause:2017} is the baseline, while AvgLen and StdDev denote the average number of words in the paragraph and standard deviation of them respectively. N, V and P are Nouns, Verbs and Pronouns observed in the paragraph.}
\label{ipg:langanal}
\end{table}

\section{Conclusion and Future Work}
\label{sec:cofw}
We have presented SGST, treating vision-to-sequence as graph-to-sequence learning. We encode images into scene graphs and condition on them for long textual sequence generation. Our experiments show that our proposed approach can effectively encode scene graphs for generating paragraphs. In future, we plan to investigate the impact of leveraging graph reasoning while encoding scene graph constituents into vectors. Further, we also aim to find the impact of sparse attention on the attention heads and compare the performance with GCN encoders.

\section{Acknowledgements}
\label{sec:ack}
This work was supported by the German Research Foundation (DFG) as a part of - Project-ID 232722074 - SFB1102.

\bibliography{example_paper}

\begin{thebibliography}{36}
\providecommand{\natexlab}[1]{#1}
\providecommand{\url}[1]{\texttt{#1}}
\expandafter\ifx\csname urlstyle\endcsname\relax
  \providecommand{\doi}[1]{doi: #1}\else
  \providecommand{\doi}{doi: \begingroup \urlstyle{rm}\Url}\fi

\bibitem[Anderson et~al.(2017)Anderson, He, Buehler, Teney, Johnson, Gould, and
  Zhang]{anderson:2017}
Anderson, P., He, X., Buehler, C., Teney, D., Johnson, M., Gould, S., and
  Zhang, L.
\newblock Bottom-up and top-down attention for image captioning and vqa.
\newblock \emph{arXiv preprint arXiv:1707.07998}, 2017.

\bibitem[Bahdanau et~al.(2014)Bahdanau, Cho, and Bengio]{bahdanau:2014}
Bahdanau, D., Cho, K., and Bengio, Y.
\newblock Neural machine translation by jointly learning to align and
  translate.
\newblock \emph{arXiv preprint arXiv:1409.0473}, 2014.

\bibitem[Beck et~al.(2018)Beck, Haffari, and Cohn]{beckgraph:2018}
Beck, D., Haffari, G., and Cohn, T.
\newblock Graph-to-sequence learning using gated graph neural networks.
\newblock \emph{arXiv preprint arXiv:1806.09835}, 2018.

\bibitem[Blondel et~al.(2019)Blondel, Martins, and Niculae]{blondel:2019}
Blondel, M., Martins, A., and Niculae, V.
\newblock Learning classifiers with fenchel-young losses: Generalized
  entropies, margins, and algorithms.
\newblock In \emph{The 22nd International Conference on Artificial Intelligence
  and Statistics}, pp.\  606--615, 2019.

\bibitem[Chatterjee \& Schwing(2018)Chatterjee and Schwing]{chatterjee:2018}
Chatterjee, M. and Schwing, A.~G.
\newblock Diverse and coherent paragraph generation from images.
\newblock In \emph{Proceedings of the European Conference on Computer Vision
  (ECCV)}, pp.\  729--744, 2018.

\bibitem[Child et~al.(2019)Child, Gray, Radford, and Sutskever]{child:2019}
Child, R., Gray, S., Radford, A., and Sutskever, I.
\newblock Generating long sequences with sparse transformers.
\newblock \emph{arXiv preprint arXiv:1904.10509}, 2019.

\bibitem[Cho et~al.(2014)Cho, Van~Merri{\"e}nboer, Gulcehre, Bahdanau,
  Bougares, Schwenk, and Bengio]{cholearning:2014}
Cho, K., Van~Merri{\"e}nboer, B., Gulcehre, C., Bahdanau, D., Bougares, F.,
  Schwenk, H., and Bengio, Y.
\newblock Learning phrase representations using rnn encoder-decoder for
  statistical machine translation.
\newblock \emph{arXiv preprint arXiv:1406.1078}, 2014.

\bibitem[Correia et~al.(2019)Correia, Niculae, and Martins]{correia:2019}
Correia, G.~M., Niculae, V., and Martins, A.~F.
\newblock Adaptively sparse transformers.
\newblock In \emph{Proceedings of the 2019 Conference on Empirical Methods in
  Natural Language Processing and the 9th International Joint Conference on
  Natural Language Processing (EMNLP-IJCNLP)}, pp.\  2174--2184, 2019.

\bibitem[Denkowski \& Lavie(2014)Denkowski and Lavie]{lavie:2014}
Denkowski, M. and Lavie, A.
\newblock Meteor universal: Language specific translation evaluation for any
  target language.
\newblock In \emph{Proceedings of the ninth workshop on statistical machine
  translation}, pp.\  376--380, 2014.

\bibitem[Johnson et~al.(2015)Johnson, Krishna, Stark, Li, Shamma, Bernstein,
  and Fei-Fei]{johnson:2015}
Johnson, J., Krishna, R., Stark, M., Li, L.-J., Shamma, D., Bernstein, M., and
  Fei-Fei, L.
\newblock Image retrieval using scene graphs.
\newblock In \emph{Proceedings of the IEEE conference on computer vision and
  pattern recognition}, pp.\  3668--3678, 2015.

\bibitem[Karpathy \& Fei-Fei(2015)Karpathy and Fei-Fei]{karpathy:2015}
Karpathy, A. and Fei-Fei, L.
\newblock Deep visual-semantic alignments for generating image descriptions.
\newblock In \emph{Proceedings of the IEEE Conference on Computer Vision and
  Pattern Recognition}, pp.\  3128--3137, 2015.

\bibitem[Kennedy et~al.(2007)Kennedy, Bugajska, Marge, Adams, Fransen,
  Perzanowski, Schultz, and Trafton]{kennedyspatial:2007}
Kennedy, W.~G., Bugajska, M.~D., Marge, M., Adams, W., Fransen, B.~R.,
  Perzanowski, D., Schultz, A.~C., and Trafton, J.~G.
\newblock Spatial representation and reasoning for human-robot collaboration.
\newblock In \emph{AAAI}, volume~7, pp.\  1554--1559, 2007.

\bibitem[Kingma \& Ba(2014)Kingma and Ba]{kingma:2014}
Kingma, D. and Ba, J.
\newblock Adam: A method for stochastic optimization.
\newblock \emph{arXiv preprint arXiv:1412.6980}, 2014.

\bibitem[Kipf \& Welling(2016)Kipf and Welling]{kipfsemi:2016}
Kipf, T.~N. and Welling, M.
\newblock Semi-supervised classification with graph convolutional networks.
\newblock \emph{arXiv preprint arXiv:1609.02907}, 2016.

\bibitem[Koncel-Kedziorski et~al.(2019)Koncel-Kedziorski, Bekal, Luan, Lapata,
  and Hajishirzi]{koncel:2019}
Koncel-Kedziorski, R., Bekal, D., Luan, Y., Lapata, M., and Hajishirzi, H.
\newblock Text generation from knowledge graphs with graph transformers.
\newblock \emph{arXiv preprint arXiv:1904.02342}, 2019.

\bibitem[Krause et~al.(2017)Krause, Johnson, Krishna, and Fei-Fei]{krause:2017}
Krause, J., Johnson, J., Krishna, R., and Fei-Fei, L.
\newblock A hierarchical approach for generating descriptive image paragraphs.
\newblock In \emph{Proceedings of the IEEE Conference on Computer Vision and
  Pattern Recognition}, pp.\  317--325, 2017.

\bibitem[Krishna et~al.(2017)Krishna, Zhu, Groth, Johnson, Hata, Kravitz, Chen,
  Kalantidis, Li, Shamma, et~al.]{krishna:2017}
Krishna, R., Zhu, Y., Groth, O., Johnson, J., Hata, K., Kravitz, J., Chen, S.,
  Kalantidis, Y., Li, L.-J., Shamma, D.~A., et~al.
\newblock Visual genome: Connecting language and vision using crowdsourced
  dense image annotations.
\newblock \emph{International Journal of Computer Vision}, 123\penalty0
  (1):\penalty0 32--73, 2017.

\bibitem[Liang et~al.(2017)Liang, Hu, Zhang, Gan, and Xing]{liang:2017}
Liang, X., Hu, Z., Zhang, H., Gan, C., and Xing, E.~P.
\newblock Recurrent topic-transition gan for visual paragraph generation.
\newblock In \emph{Proceedings of the IEEE International Conference on Computer
  Vision}, pp.\  3362--3371, 2017.

\bibitem[Lin et~al.(2014)Lin, Maire, Belongie, Hays, Perona, Ramanan,
  Doll{\'a}r, and Zitnick]{lin:2014}
Lin, T.-Y., Maire, M., Belongie, S., Hays, J., Perona, P., Ramanan, D.,
  Doll{\'a}r, P., and Zitnick, C.~L.
\newblock Microsoft coco: Common objects in context.
\newblock In \emph{European conference on computer vision}, pp.\  740--755.
  Springer, 2014.

\bibitem[Martins \& Astudillo(2016)Martins and Astudillo]{martins:2016}
Martins, A. and Astudillo, R.
\newblock From softmax to sparsemax: A sparse model of attention and
  multi-label classification.
\newblock In \emph{International Conference on Machine Learning}, pp.\
  1614--1623, 2016.

\bibitem[Papineni et~al.(2002)Papineni, Roukos, Ward, and Zhu]{papineni:2002}
Papineni, K., Roukos, S., Ward, T., and Zhu, W.-J.
\newblock Bleu: a method for automatic evaluation of machine translation.
\newblock In \emph{Proceedings of the 40th annual meeting on association for
  computational linguistics}, pp.\  311--318. Association for Computational
  Linguistics, 2002.

\bibitem[Peters et~al.(2019)Peters, Niculae, and Martins]{peters:2019}
Peters, B., Niculae, V., and Martins, A.~F.
\newblock Sparse sequence-to-sequence models.
\newblock \emph{arXiv preprint arXiv:1905.05702}, 2019.

\bibitem[Radford et~al.(2018)Radford, Narasimhan, Salimans, and
  Sutskever]{radford:2018}
Radford, A., Narasimhan, K., Salimans, T., and Sutskever, I.
\newblock Improving language understanding by generative pre-training.
\newblock \emph{URL https://s3-us-west-2. amazonaws.
  com/openai-assets/researchcovers/languageunsupervised/language understanding
  paper. pdf}, 2018.

\bibitem[Ren et~al.(2015)Ren, He, Girshick, and Sun]{ren:2015}
Ren, S., He, K., Girshick, R., and Sun, J.
\newblock Faster r-cnn: Towards real-time object detection with region proposal
  networks.
\newblock In \emph{Advances in neural information processing systems}, pp.\
  91--99, 2015.

\bibitem[Rennie et~al.(2016)Rennie, Marcheret, Mroueh, Ross, and
  Goel]{rennie:2016}
Rennie, S.~J., Marcheret, E., Mroueh, Y., Ross, J., and Goel, V.
\newblock Self-critical sequence training for image captioning.
\newblock \emph{arXiv preprint arXiv:1612.00563}, 2016.

\bibitem[Song et~al.(2018)Song, Zhang, Wang, and Gildea]{song:2018}
Song, L., Zhang, Y., Wang, Z., and Gildea, D.
\newblock A graph-to-sequence model for amr-to-text generation.
\newblock \emph{arXiv preprint arXiv:1805.02473}, 2018.

\bibitem[Vaswani et~al.(2017)Vaswani, Shazeer, Parmar, Uszkoreit, Jones, Gomez,
  Kaiser, and Polosukhin]{vaswani:2017}
Vaswani, A., Shazeer, N., Parmar, N., Uszkoreit, J., Jones, L., Gomez, A.~N.,
  Kaiser, {\L}., and Polosukhin, I.
\newblock Attention is all you need.
\newblock In \emph{Advances in neural information processing systems}, pp.\
  5998--6008, 2017.

\bibitem[Vedantam et~al.(2015)Vedantam, Lawrence~Zitnick, and
  Parikh]{vedantam:2015}
Vedantam, R., Lawrence~Zitnick, C., and Parikh, D.
\newblock Cider: Consensus-based image description evaluation.
\newblock In \emph{Proceedings of the IEEE conference on computer vision and
  pattern recognition}, pp.\  4566--4575, 2015.

\bibitem[Wang et~al.(2019)Wang, Pan, Yao, Tang, and
  Mei]{wangconvolutional:2019}
Wang, J., Pan, Y., Yao, T., Tang, J., and Mei, T.
\newblock Convolutional auto-encoding of sentence topics for image paragraph
  generation.
\newblock \emph{IJCAI}, 2019.

\bibitem[Xu et~al.(2015)Xu, Ba, Kiros, Cho, Courville, Salakhudinov, Zemel, and
  Bengio]{xu:2015}
Xu, K., Ba, J., Kiros, R., Cho, K., Courville, A., Salakhudinov, R., Zemel, R.,
  and Bengio, Y.
\newblock Show, attend and tell: Neural image caption generation with visual
  attention.
\newblock In \emph{International Conference on Machine Learning}, pp.\
  2048--2057, 2015.

\bibitem[Yang et~al.(2019{\natexlab{a}})Yang, Tang, Zhang, and
  Cai]{yangauto:2019}
Yang, X., Tang, K., Zhang, H., and Cai, J.
\newblock Auto-encoding scene graphs for image captioning.
\newblock In \emph{Proceedings of the IEEE Conference on Computer Vision and
  Pattern Recognition}, pp.\  10685--10694, 2019{\natexlab{a}}.

\bibitem[Yang et~al.(2019{\natexlab{b}})Yang, Dai, Yang, Carbonell,
  Salakhutdinov, and Le]{yangxlnet:2019}
Yang, Z., Dai, Z., Yang, Y., Carbonell, J., Salakhutdinov, R., and Le, Q.~V.
\newblock Xlnet: Generalized autoregressive pretraining for language
  understanding.
\newblock \emph{arXiv preprint arXiv:1906.08237}, 2019{\natexlab{b}}.

\bibitem[Yao et~al.(2018)Yao, Pan, Li, and Mei]{yaoexploring:2018}
Yao, T., Pan, Y., Li, Y., and Mei, T.
\newblock Exploring visual relationship for image captioning.
\newblock In \emph{Proceedings of the European Conference on Computer Vision
  (ECCV)}, pp.\  684--699, 2018.

\bibitem[Yun et~al.(2019)Yun, Jeong, Kim, Kang, and Kim]{yungraph:2019}
Yun, S., Jeong, M., Kim, R., Kang, J., and Kim, H.~J.
\newblock Graph transformer networks.
\newblock In \emph{Advances in Neural Information Processing Systems}, pp.\
  11960--11970, 2019.

\bibitem[Zellers et~al.(2018)Zellers, Yatskar, Thomson, and Choi]{zellers:2018}
Zellers, R., Yatskar, M., Thomson, S., and Choi, Y.
\newblock Neural motifs: Scene graph parsing with global context.
\newblock In \emph{Proceedings of the IEEE Conference on Computer Vision and
  Pattern Recognition}, pp.\  5831--5840, 2018.

\bibitem[Zhu et~al.(2019)Zhu, Li, Zhu, Qian, Zhang, and Zhou]{zhugraph:2019}
Zhu, J., Li, J., Zhu, M., Qian, L., Zhang, M., and Zhou, G.
\newblock Modeling graph structure in transformer for better amr-to-text
  generation.
\newblock In \emph{Proceedings of the 2019 Conference on Empirical Methods in
  Natural Language Processing and the 9th International Joint Conference on
  Natural Language Processing (EMNLP-IJCNLP)}, pp.\  5462--5471, 2019.

\end{thebibliography}
\bibliographystyle{icml2020}

\end{document}